%% file: main.tex
\documentclass[runningheads]{llncs}
\usepackage[T1]{fontenc}
\usepackage{graphicx}

\usepackage{amsmath}
\usepackage{color}
\usepackage{hyperref}
\usepackage{multirow}
\usepackage{tikz}
\usepackage{float}
\usetikzlibrary{arrows.meta, positioning, shapes.geometric}

\begin{document}

\title{Imitation of Arm Gestures by the Semi-Humanoid Robot NICO}
\titlerunning{Arm Gesture Imitation by Robot NICO}
\author{Anastasiya Ihnatovich, Igor Farkaš} 
\institute{Faculty of Mathematics, Physics and Informatics\\
Comenius University Bratislava\\
Mlynská dolina 6284, 842 48 Bratislava, Slovakia}
\maketitle

\begin{abstract}
Seamless human--robot interaction (HRI) requires a number of perceptual and motor abilities from the robot, one of them being the imitation of human gestures. Humanoid robots have an advantage in HRI thanks to their anthropomorphic features.
In this work, we develop a system for imitation of human arm gestures by the semi-humanoid robot NICO based on analytical geometry and a pretrained MediaPipe pose-estimation model. 
For each input RGB frame, 3D coordinates of relevant human body landmarks, including arm joints and hand keypoints, are obtained using the MediaPipe framework. Joint angles are then computed from these coordinates using derived geometric relations. Finally, the computed angles are properly mapped to NICO’s motor configuration and executed in a predefined motion sequence. Preliminary experiments on several representative arm gestures with six participants of different height indicate that the proposed method can produce meaningful imitative motions from monocular RGB input only, while also highlighting limitations in more complex poses and wrist-related movements.

\keywords{gesture imitation \and joint angle reconstruction \and angle mapping \and MediaPipe accuracy.}
\end{abstract}

\section{Introduction}

Robotic imitation is a fundamental capability for natural human–robot interaction, enabling robots to acquire skills through observation rather than explicit programming. Early approaches relied on direct trajectory teaching, where a human physically guided the robot while recording joint configurations. This paradigm evolved with the introduction of inverse kinematics (IK) methods, which compute joint parameters required to reproduce desired motions. More recently, learning-based approaches have been proposed, framing imitation as a data-driven problem. Foundational work highlights imitation as a key pathway toward humanoid intelligence \cite{schaal99} while recent studies emphasize multimodal perception and embodied learning \cite{spisak24}. Neural network approaches based on self-supervised learning \cite{lucny23} and pose similarity metric learning \cite{lei14} demonstrate strong generalization capabilities but typically require large datasets, extensive training, and significant computational resources, which can limit real-time applicability.
In contrast, analytical and geometry-based approaches offer a computationally efficient and interpretable alternative (e.g. \cite{handMediapipe,depthGeometry}), albeit with some limitations (see Related work). 

The emergence of RGB-based pose estimation frameworks has turned out to be quite successful. MediaPipe \cite{mediapipe} enables real-time extraction of 2D and pseudo-3D body and hand landmarks from monocular images, making it suitable for low-cost systems. Its accuracy has been evaluated \cite{mediapipeAccuracy}, where results show that joint angle estimation is generally reliable, although performance degrades under occlusions and challenging viewpoints. These findings suggest that RGB-only pipelines are viable but require careful handling of noise and missing data.
Vision-based human–machine interaction has also been explored in the context of robotic hand control. It has been demonstrated that human hand movements can be used to control robotic hands through vision-based tracking, enabling intuitive interaction \cite{garcia25}. However, such approaches typically focus on hand articulation and do not address full arm motion or coordinated gesture imitation.

Overall, existing work demonstrates that imitation can be achieved using both analytical and learning-based approaches. Learning-based methods offer flexibility and generalization but at the cost of complexity, while analytical approaches provide efficiency and transparency but often require adaptation to specific kinematic structures. Moreover, prior studies either focus on simplified gestures or depend on specialized sensing hardware. 

In this context, the present work extends analytical geometry-based methods to a broader set of arm gestures using only RGB input. By leveraging MediaPipe for landmark extraction and adapting geometric formulations to the kinematics of the NICO robot, the proposed approach achieves meaningful imitation without reliance on expensive sensors or computationally intensive learning models.




\section{Related Work}

With the advancement of computer vision, systems have emerged for extracting human skeletons from RGB images. One of the most well-known such systems is OpenPose~\cite{openpose}, which is designed for multi-user 2D pose estimation, using two CNNs to predict body parts and their associations.

More recent solutions, such as MediaPipe, enable the real-time extraction of pseudo-3D landmarks using standard RGB cameras~\cite{mediapipe} . Thanks to its low latency, MediaPipe has become a popular solution for human-robot interaction and gesture-based systems.

With the emergence of pose-estimation systems, vision-based human motion imitation became more feasible. Instead of relying on wearable sensors or motion-capture systems, robots can now infer human body configuration directly from RGB images and reproduce the observed motion. This development motivated both analytical geometry-based methods and learning-based imitation approaches.

Early neural-network-based approaches to robot imitation were explored by Durdu et al., who used artificial neural networks to classify human arm movements collected from joint-mounted sensors and reproduce them on a robotic arm platform~\cite{imitation_ann}.

Self-supervised approaches were explored by Lúčny et al.\cite{lucny23}, where robots learn imitation through associations between visual observations and their own motor states. These methods reduce the need for manually annotated datasets and demonstrate promising generalization capabilities. 

Although learning-based methods demonstrate high flexibility and the ability to generalize, they typically require large-scale datasets, training procedures, and substantial computational resources.

A number of studies use analytical geometry to transform landmark coordinates into joint angles without relying on learning-based methods. Altayeb \cite{handMediapipe} used MediaPipe to extract hand landmarks, while joint angles were computed using direct geometric relations for further robotic hand control. Their study demonstrates that analytical methods can achieve high accuracy without relying on learning-based approaches. The reported accuracy of approximately 96\% indicates that analytical approaches are promising, but their applicability to more complex movements remains an open question.


Another relevant study \cite{depthGeometry} presents a system based on depth-camera input. In this work, joint angles are derived using geometric relations and mapped to the NAO robot platform. The authors report an imitation accuracy of approximately 95\% for arm motion imitation, demonstrating that analytical methods can be effective when accurate 3D data is available. However, the reliance on depth-sensing hardware increases system cost and complexity.

In contrast, the present work applies a similar analytical framework to the humanoid robot NICO using only RGB camera input. Since the kinematic structure of NICO differs from that of the NAO robot, the corresponding angle formulations must also be adapted. In particular, the proposed system includes estimation of forearm rotation and wrist flexion, while NICO does not provide the elbow roll joint available on the NAO platform. Furthermore, this work incorporates MediaPipe hand landmarks, which provide additional information for estimating wrist flexion  and forearm rotation. Head motion imitation is outside the scope of the current implementation and is therefore not addressed experimentally.

Overall, existing analytical approaches demonstrate that promising pose imitation is achievable without learning-based models. However, prior work either focuses on simplified gesture sets or depends on specialized hardware. 

Thus, the proposed approach does not require a training stage, uses only monocular RGB input, supports full arm and hand reconstruction, and is tailored to the kinematics of the NICO robot.

\section{Materials and Methods}

Here we describe the hardware and software used, as well as the processing pipeline for gesture imitation.

\subsection{NICO Robot}
\label{sec:nicoDescription}

\begin{figure}[h]
    \centering
    \includegraphics[width=0.6\textwidth]{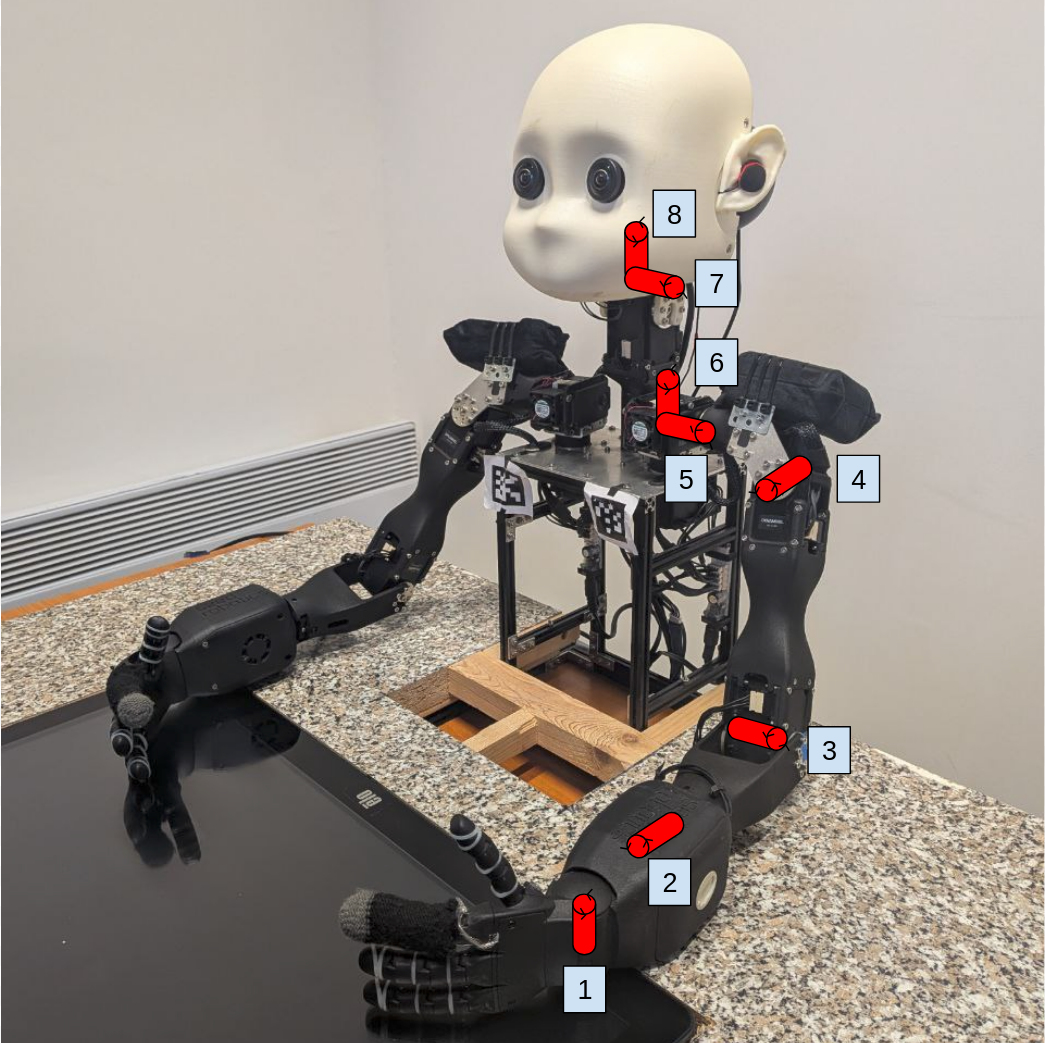}
    \caption{Degrees of freedom of the NICO robot.}
    \label{fig:nico-dofs}
\end{figure}

NICO (Neuro-Inspired COmpanion) \cite{nicoPlatform} is a child-size semi-humanoid robot developed for social interaction by the Knowledge Technology group at the University of Hamburg, shown in Fig.~\ref{fig:nico-dofs}. The robot is equipped with two actuated degrees of freedom (DoF) in the head and six DoFs in each arm, enabling a wide range of upper-body motions suitable for gesture imitation tasks. The joints are driven by Dynamixel servo actuators, which provide programmable position control together with feedback such as present position and load estimates.

\noindent Now we describe the processing pipeline used for gesture imitation. The proposed framework converts monocular RGB input into robot motion through several consecutive stages. Human pose and hand landmarks are first estimated using MediaPipe. These landmarks are then processed by analytical geometric formulas to recover the required joint angles. In the final stage, the obtained angles are transformed into valid motor commands for the NICO robot. An overview of the complete pipeline is shown in Fig.~\ref{fig:pipeline}.

\input{figures/pipeline_diagram}

\subsection{3D Coordinates Extraction}

A key limitation of the proposed system is the use of a single RGB camera without additional sensing modalities such as depth sensors. To estimate 3D human pose under this constraint, the MediaPipe framework~\cite{mediapipe} is employed. MediaPipe provides models for extracting both pose and hand landmarks directly from RGB images.

\begin{figure}[h]
  \centering
  \includegraphics[width=0.8\textwidth]{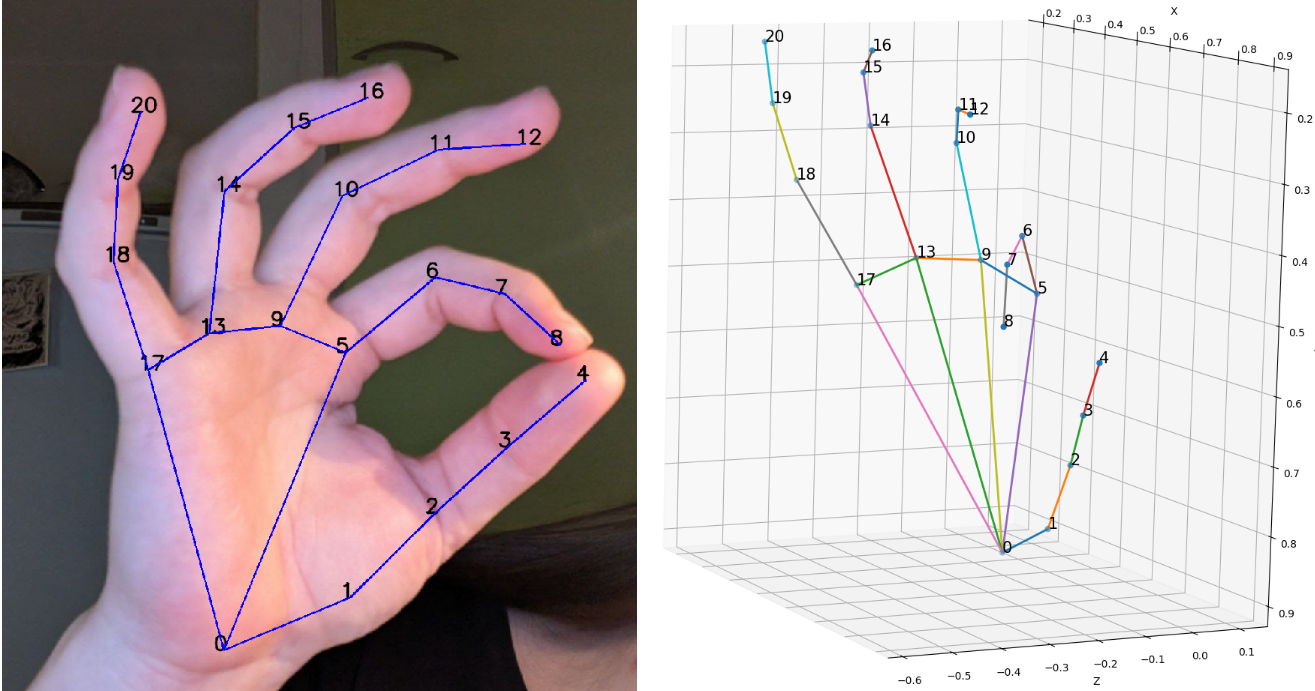}
  \caption{Visualization of hand landmarks extracted using MediaPipe.}
  \label{fig:mediapipe-hand}
\end{figure}

\begin{figure}
  \centering
  \includegraphics[width=1\textwidth]{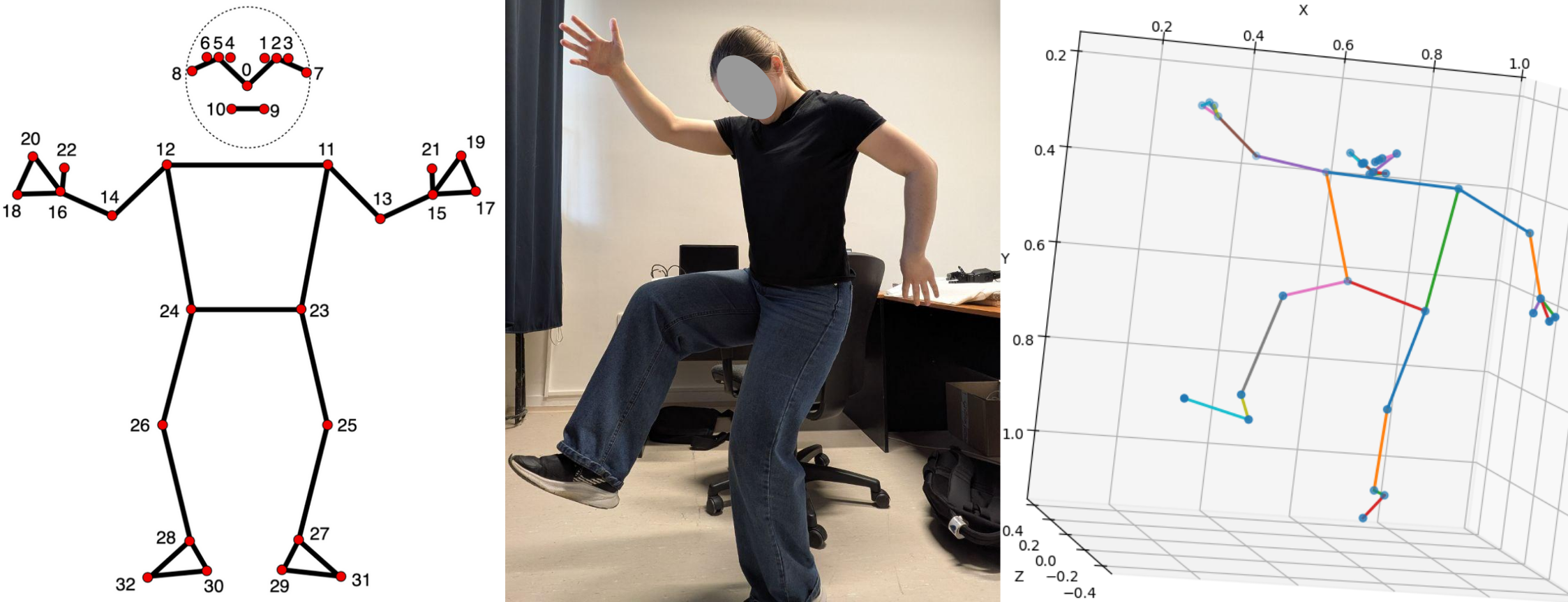}
  \caption{Example of pose landmark extraction. {\it Left:} MediaPipe landmark indexing scheme. {\it Center:} input RGB frame. {\it Right:} reconstructed 3D landmarks obtained from MediaPipe.}
  \label{fig:mediapipe-pose-pattern}
\end{figure}

\noindent The hand landmark model outputs 3D coordinates of 21 key points, as illustrated in Fig.~\ref{fig:mediapipe-hand}. The $x$ and $y$ coordinates are normalized with respect to the image dimensions, while the $z$ coordinate is defined relative to the wrist (landmark 0).
For full-body pose estimation, MediaPipe detects 33 landmarks corresponding to major body joints (Fig.~\ref{fig:mediapipe-pose-pattern}). We used the \texttt{pose\_landmarker\_heavy.task} model, which provides higher accuracy than the lite and full variants at the cost of increased computational demands. Each landmark approximates the position of a specific body part (e.g., eyes, shoulders, elbows). The $z$ coordinate is defined relative to the estimated depth of the hips.

Whenever hand landmarks were available, the corresponding wrist and palm points from the hand detector were preferred over pose landmarks. Specifically, pose landmarks $P_{15}, P_{17}, P_{19}$ for the left hand and $P_{16}, P_{18}, P_{20}$ for the right hand were replaced by the corresponding hand-model landmarks. This substitution was used for palm orientation, wrist flexion, and forearm rotation. Since the hand landmarks are expressed in a local coordinate system centered at the hand, the landmark set was translated so that its origin ($P_0$) coincided with the corresponding wrist landmark in the pose model. No additional scaling or rotational alignment was required, as both MediaPipe Pose and MediaPipe Hands estimate landmarks within the same camera coordinate frame, differing only in the choice of origin.

\noindent In this work, we use landmarks 11--24, focusing on arm motion. The approach can be extended to full-body imitation, which remains an interesting direction for future investigation.

The accuracy of MediaPipe outputs has been evaluated in prior work~\cite{mediapipeAccuracy}. In that study, four subjects performed multiple physical exercises recorded from different viewpoints, with ground truth provided by a motion capture system (MTw Awinda). The results indicate that estimation accuracy strongly depends on the viewing angle, with increased error under occlusions. Additionally, joint angle estimation was found to be more reliable than absolute body measurements.

These findings are consistent with observations in this work, where occlusions and non-optimal viewpoints lead to decreased precision.

\subsection{Joint Angle Reconstruction}
To save space, the calculations are described for the left arm. The calculations for the right arm are performed in the same way, taking into account the orientation of the rotations when taking the vector product. 

The following formulas are derived assuming that the subject's torso is approximately parallel to the camera plane. To compensate for deviations from this assumption, the landmark coordinates are rotated about the vertical (yaw) axis using the orientation of the shoulder line estimated from the left and right shoulder landmarks. This alignment reduces the effect of torso yaw on the reconstructed joint angles. Compensation for torso pitch and roll is not performed, as the experimental protocol considers an upright standing posture in which these rotations are expected to be small.

\begin{table}[h]
\centering
\caption{Description of the symbols used.}
\label{tab:notation}
\begin{tabular}{p{2.2cm} p{10cm}}
\hline
Symbol & Explanation \\
\hline

$P_i$ &
3D point of extracted landmark $i$ (using the numbering from Fig.~\ref{fig:mediapipe-pose-pattern}). \\

$V_{\rm LA}$ &
Vector of left upper arm, pointing from shoulder ($P_{11}$) to elbow ($P_{13}$). \\

$V_{\rm LF}$ &
Vector of the left forearm, pointing from elbow ($P_{13}$) to wrist ($P_{15}$). \\

$V_{\rm LPN}$ &
Normal vector to the left palm plane. \\

$V_{\rm LRS}$ &
Vector of shoulders, pointing from left ($P_{11}$) to right shoulder ($P_{12}$). \\
\hline
\end{tabular}
\end{table}

Table~\ref{tab:notation} summarizes the symbols and notation used in the proposed method. The left palm-plane normal vector is defined as
\begin{equation}
V_{\rm LPN} = (P_{17}-P_{15}) \times (P_{19}-P_{15}).
\end{equation}

\subsubsection{Wrist flexion angle}
(DoF 1 in Fig.~\ref{fig:nico-dofs}) is given by
\begin{equation}
\theta_{\mathrm{\text{wrist-bend}}} = \frac{\pi}{2} + \angle A,
\end{equation}
where $\angle A$ denotes unsigned angle between $V_{\rm LPN}$ and $(-V_{\rm LF})$.
Geometrically, this follows from the relationship between the forearm direction and the palm plane normal. The same formula applies for both downward and upward bending, ensuring a continuous definition over the full range of motion.

\subsubsection{Forearm rotation angle}
(DoF 2) can be computed as follows:

\paragraph{Reference direction.} The reference rotation base is defined as the component of $V_{\rm LA}$ orthogonal to $V_{\rm LF}$:
\begin{equation}
rot. base = V_{\rm LA} - \frac{V_{\rm LA} \cdot V_{\rm LF}}{V_{\rm LF} \cdot V_{\rm LF}} V_{\rm LF}.
\end{equation}

\begin{figure}[h!]
  \begin{center}
     \includegraphics[width=1\textwidth]{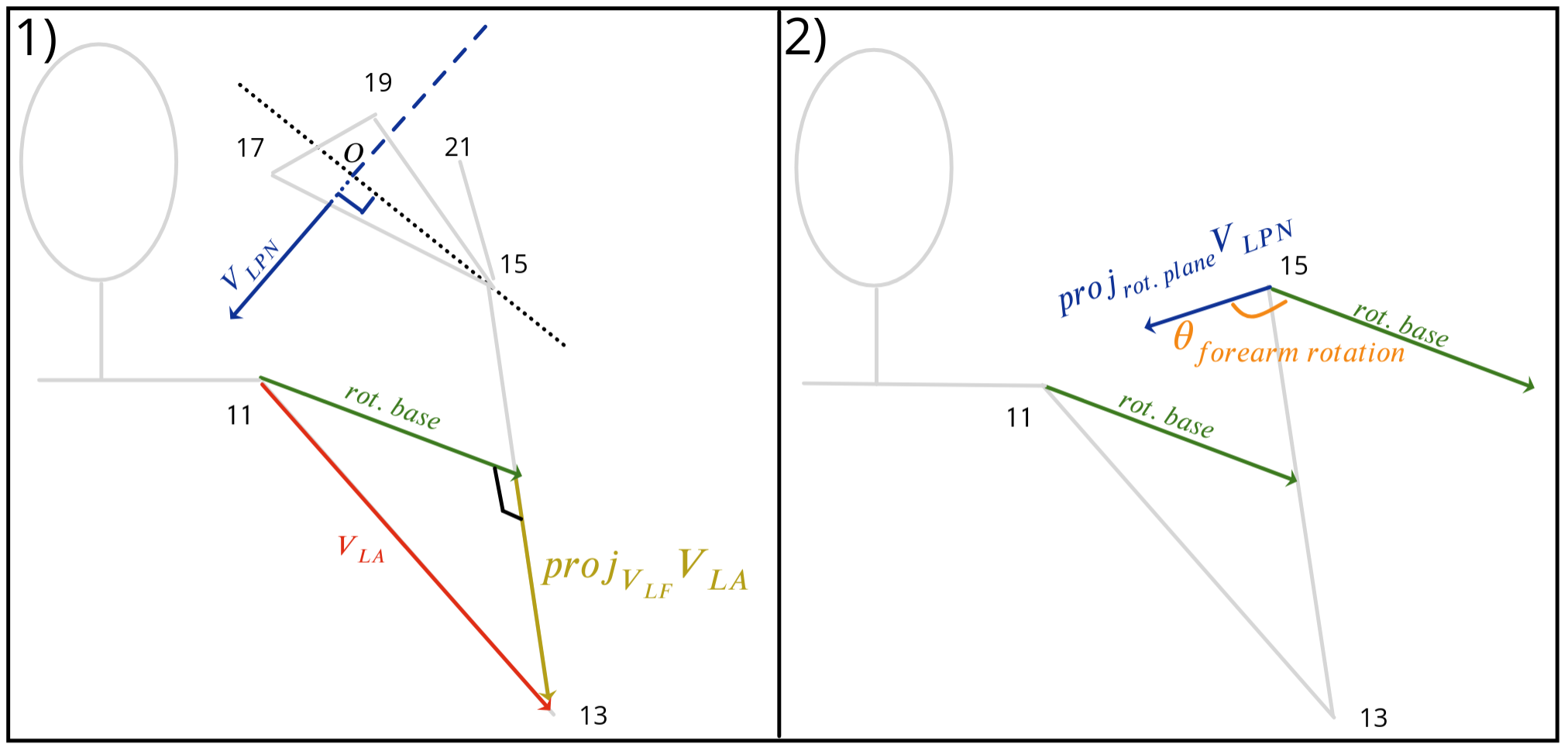} \caption{Example of forearm rotation calculations. Numeration from Fig.~\ref{fig:mediapipe-pose-pattern}}
     \label{fig:forearm-rotation}
  \end{center}
\end{figure}

\noindent In case of $\|rot.base\| = 0$ (straight arm), a different method of calculation is needed. In this case, NICO's joints design allows us to calculate the reference direction as
\begin{equation}
rot.base = V_{\rm LA} \times V_{\rm LRS},  
\end{equation}
where the symbol $\times$ denotes the cross product of two vectors. The resulting vector is multiplied by $-1$ when the z-coordinate of $V_{\rm LA}$ is positive, since in this configuration the effect of the arm being behind the body is accounted for by the shoulder roll calculation rather than the shoulder pitch calculation (see Shoulder angles below).

When $V_{\rm LA}$ and $V_{\rm LRS}$ are parallel, fallback vector $(0,1,0)^\top$ is used.

\paragraph{Palm orientation.}
Let $proj_{rot. plane} (V_{\rm LPN})$ be the projection of the palm normal onto rotation plane, i.e. the plane perpendicular to $V_{\rm LF}$. It is calculated as
\begin{equation}
proj_{rot. plane}(V_{\rm LPN}) = V_{\rm LPN} - \frac{V_{\rm LPN} \cdot V_{\rm LF}}{V_{\rm LF} \cdot V_{\rm LF}} V_{\rm LF}.
\end{equation}
If $\|proj_{rot. plane}(V_{\rm LPN}) \| \neq 0$, the palm orientation vector is defined as
\begin{equation} 
V_{\rm PO} = -\text{sgn}(\cos (\theta_{\text{wrist-bend}})) \cdot proj_{rot. plane}(V_{\rm LPN}). 
\end{equation}
Otherwise (when palm is perpendicular to forearm), a fallback vector is constructed using the wrist point and the midpoint of landmarks $P_{17}$ and $P_{19}$:
\begin{equation}
M = 0.5(P_{17}+P_{19}), \qquad V_{W} = M-P_{15},
\end{equation}
and the palm orientation vector is defined as
\begin{equation} 
V_{\rm PO} = \text{sgn}(\pi- \theta_{\text{wrist-bend}}) \cdot V_{\rm W} \end{equation}

\paragraph{Forearm rotation angle}
is then given by the signed angle between palm orientation and rotation base vectors around the forearm axis.  An illustrative example is shown in Fig.~\ref{fig:forearm-rotation}.

\subsubsection{Elbow bend angle}

To compute elbow bend angle (DoF 3) in range $(0, \pi]$ it is enough to compute unsigned angle between forearm and upper arm vectors
\begin{equation}
\theta_{\text{elbow-bend}} =
\arccos\!\left(
\frac{V_{\rm LF} \cdot V_{\rm LA}}
{\|V_{\rm LF}\| \cdot \|V_{\rm LA}\|}
\right)
\end{equation}

\subsubsection{Shoulder angles}

\begin{figure}
  \begin{center}
     \includegraphics[width=1\textwidth]{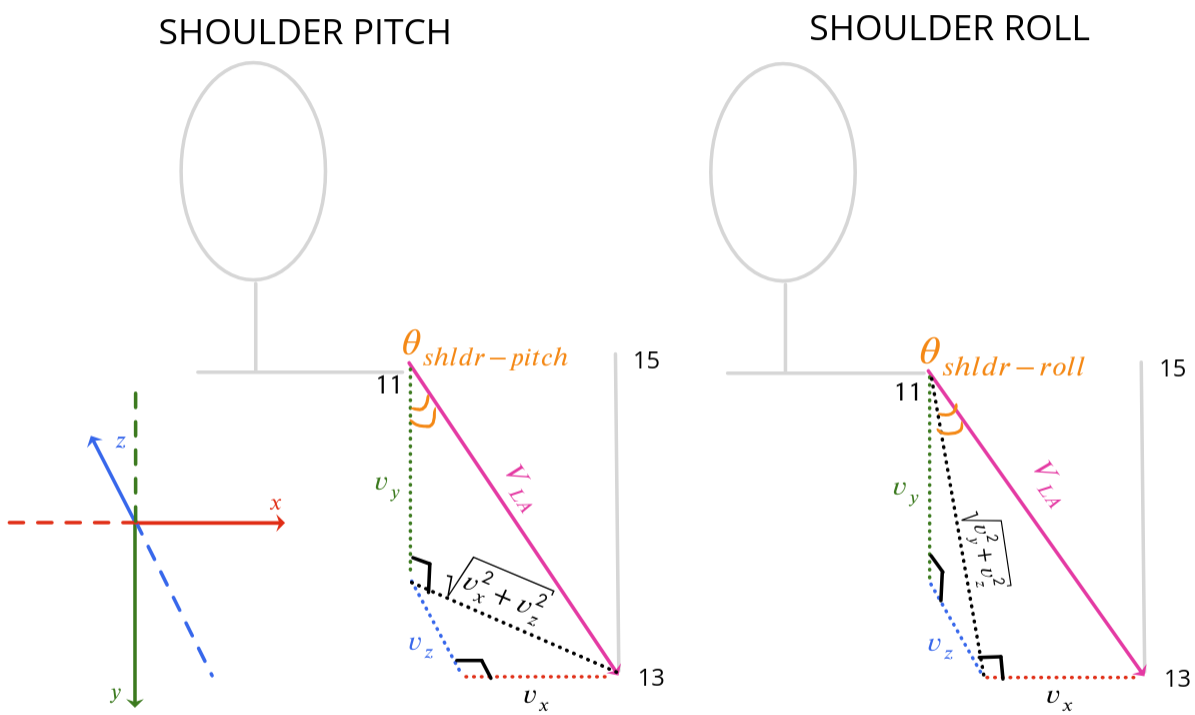} \caption{Left shoulder angles calculations. Numeration from Fig.~\ref{fig:mediapipe-pose-pattern}}
     \label{fig:shoulder-angles}
  \end{center}
\end{figure}

\paragraph{Shoulder pitch}
To compute the left $\theta_{\text{shldr-pitch}}$ angle (DoF 5), the vector $V_{\rm LA}$ is decomposed into its vertical component and its projection onto the $O_{xz}$ plane. As seen in Fig.~\ref{fig:shoulder-angles}, the desired angle can be derived from a right triangle with sides of length $v_y$ and $(v_x^2+v_z^2)^{1/2}\geq 0$. 
To distinguish between cases of bending the arm
below and above the shoulder, the function $\text{atan2}$ automatically resolves the correct quadrant based on the sign of $v_y$:
\begin{equation}
    \theta_{\text{shldr-pitch}} = \operatorname{atan2}\!\left((v_x^2+v_z^2)^{1/2},\, v_y\right).
\end{equation}
Thus, the computed angle lies in the range $[0,\pi]$, where $0$ corresponds to the arm pointing downwards, $\pi/2$ to the arm being approximately horizontal, and $\pi$ to the arm pointing upwards. Since the term $(v_x^2+v_z^2)^{1/2}$ is non-negative, this angle does not distinguish whether the arm is in front of or behind the body. This ambiguity is handled separately in the shoulder roll computation.

\paragraph{Shoulder roll} To compute the left $\theta_{\text{shldr-roll}}$ (DoF 4), $V_{\rm LA}$ is decomposed into its horizontal component and its projection onto the $O_{yz}$ plane. As seen in Fig.~\ref{fig:shoulder-angles}, desired angle can be derived from a right triangle with sides of length $v_x$ and $(v_z^2+v_y^2)^{1/2}\geq 0$. To distinguish between cases of rotating the arm clockwise and counterclockwise from the shoulder, the function $\text{atan2}$ automatically resolves the correct quadrant based on the sign of $v_x$:
\begin{equation}
    \theta_{\text{shldr-roll}} = \operatorname{atan2}\!\left(v_x, (v_z^2+v_y^2)^{1/2}\right) + \pi/2.
\end{equation} 
For configurations in which the arm is located behind the body, identified by $v_z>0$, the computed value is mirrored with respect to $2\pi$:
\begin{equation}
\theta_{\text{shldr-roll}} = 2\pi - \theta_{\text{shldr-roll}}.
\end{equation}
This allows the method to distinguish between front and back arm configurations. The resulting angle lies in the range $[0,2\pi)$, where $\pi/2$ and $3\pi/2$ correspond to the arm being aligned with the $O_{yz}$ plane in front of and behind the body, respectively.

\paragraph{Shoulder yaw}
In the current method, shoulder yaw (DoF 6) is fixed in the neutral position.

\subsection{Angle Mapping to NICO}
\label{sec:angle-mapping}

The joint angles reconstructed from the geometric model cannot be used directly as motor commands because each NICO joint has its own admissible range and motor coordinate system. Therefore, each analytically reconstructed angle was linearly mapped to the corresponding motor range.

For each joint, the admissible analytical interval $[\theta_{\min}, \theta_{\max}]$ and the corresponding motor positions $m(\theta_{\min})$ and $m(\theta_{\max})$ were determined experimentally. The reconstructed angle $\theta$ was first clamped to the analytical interval and then mapped as
\begin{equation}
m = m(\theta_{\min}) +
\frac{\theta-\theta_{\min}}
{\theta_{\max}-\theta_{\min}}
\left(
m(\theta_{\max})-m(\theta_{\min})
\right).
\end{equation}

Since the motor positions at the interval boundaries are specified explicitly, the same equation naturally handles both matching and reversed motor orientations. Finally, the computed motor position is rounded to the nearest integer before being sent to the controller.

\section{Experiments}
The RGB-based evaluation was conducted using a dedicated dataset collected for this study. Six volunteers with different body heights participated in the experiment, including three male participants (175 cm, 183 cm, and 188 cm) and three female participants (158 cm, 168 cm, and 180 cm). 

Each participant performed 11 predefined symmetric upper-limb poses representing different combinations of shoulder pitch, shoulder roll, elbow flexion, forearm rotation, and wrist flexion (see Table ~\ref{tab:reference_poses}). Since all poses were symmetric, identical reference joint angles were used for both the left and right arm.

\begin{table}[h!]
\centering
\setlength{\tabcolsep}{9pt}
\caption{Reference joint angles (degrees) defining the evaluated poses.}
\label{tab:reference_poses}
\begin{tabular}{lccccc}
\hline
\textbf{Shoulder} &
\textbf{Shoulder} &
\textbf{Elbow} &
\textbf{Forearm} &
\textbf{Wrist} \\

\textbf{Pitch} &
\textbf{Roll} &
\textbf{Bend} &
\textbf{Rotation} &
\textbf{Bend} \\
\hline
90 & 180 & 180 & 180 & 180 \\
 45 & 135 & 180 & 45  & 180 \\
 90 & 180 & 90  & 90  & 180 \\
90 & 135 & 90  & 90  & 180 \\
45 & 180 & 45  & 180 & 180 \\
90 & 135 & 180 & 0   & 180 \\
90 & 135 & 180 & 90  & 180 \\
90 & 135 & 180 & 180 & 180 \\
45 & 180 & 180 & 0   & 240 \\
45 & 180 & 180 & 0   & 100 \\
 90 & 180 & 135 & 180 & 180 \\
\hline
\end{tabular}
\end{table}

To evaluate the influence of viewpoint, every pose was recorded under three body orientations with respect to the camera: $0^{\circ}$, $30^{\circ}$, and $45^{\circ}$. The $0^{\circ}$ configuration corresponds to a frontal view, while the remaining orientations introduce increasing self-occlusion and perspective distortion, making landmark estimation progressively more challenging.

All recordings were acquired under identical conditions. A Google Pixel 9a front-facing RGB camera was mounted at a height of approximately 125 cm above the floor and positioned 150 cm from the participant. The recording environment, camera position, illumination, and background remained unchanged throughout the experiment. To minimize the influence of clothing appearance on landmark detection, all participants wore plain black T-shirts without visible patterns.

For each captured image, body pose and hand landmarks were extracted using MediaPipe and processed by the proposed reconstruction algorithm. The reconstructed joint angles were compared with the predefined reference angles of the corresponding pose. Reconstruction accuracy was quantified using the absolute angular error for each joint. For cyclic rotational quantities, such as forearm rotation, the minimum circular angular difference was used.


The experiments demonstrate that the proposed landmark-based geometric approach provides moderate reconstruction accuracy for shoulder pitch, elbow bend, and wrist bend, achieving mean absolute errors (MAE) of approximately $10^{\circ}$, $20^{\circ}$, and $23^{\circ}$, respectively. Shoulder roll is reconstructed with lower accuracy ($34^{\circ}$ MAE), indicating that the current geometric formulation remains sensitive to body orientation and could benefit from additional normalization or geometric constraints. 

\begin{figure}
  \begin{center}
     \includegraphics[width=0.8\textwidth]{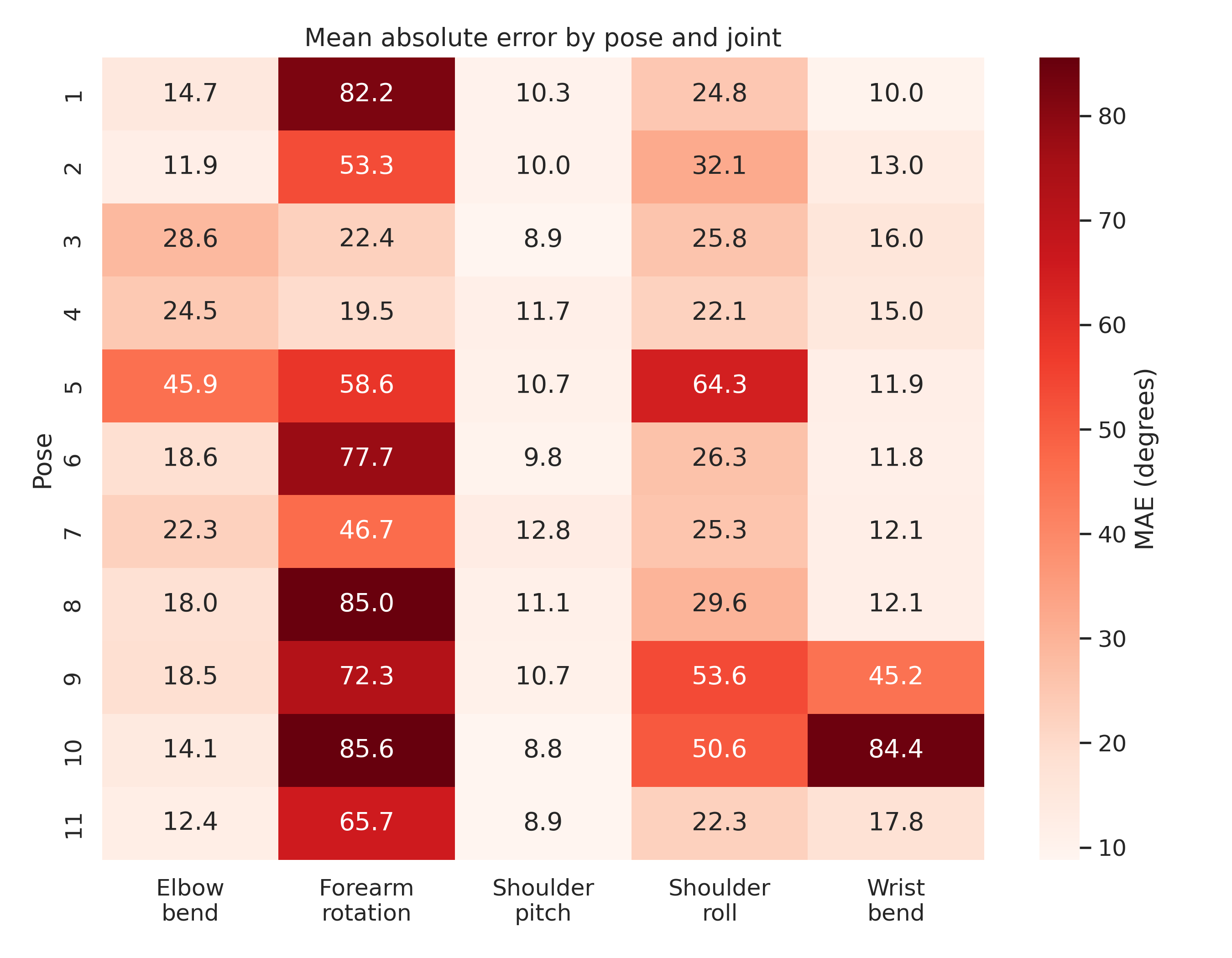} \caption{Average absolute reconstruction error (MAE, degrees) for each evaluated pose and reconstructed joint. Values are averaged across all participants and viewing angles using the proposed method with MediaPipe hand landmarks whenever available.}
     \label{fig:pose_heatmap}
  \end{center}
\end{figure}

To better illustrate how reconstruction accuracy depends on the performed gesture, Fig.~\ref{fig:pose_heatmap} presents the average absolute error for each evaluated pose and reconstructed joint. The values correspond to the proposed method using MediaPipe hand landmarks whenever available (64\% of cases) and are averaged across all participants and viewing angles.

The principal limitation of the proposed method is the estimation of forearm rotation. Using pose landmarks alone resulted in the largest reconstruction error ($67.1^{\circ}$ MAE). Replacing the forearm estimation with dedicated hand landmarks reduced the error to $50.1^{\circ}$, corresponding to an improvement of approximately $25\%$. Despite this improvement, the remaining error is still substantially higher than for the other reconstructed joints, suggesting that reliable forearm rotation estimation cannot be achieved from the available landmark configuration alone.

The limited accuracy of forearm rotation also affects wrist bend estimation. As the wrist deviates further from its neutral position, the apparent wrist angle becomes increasingly dependent on forearm orientation. Consequently, wrist bend is reconstructed considerably more accurately when the wrist remains close to its neutral configuration ($180^{\circ}$), whereas the error increases noticeably for more extreme wrist flexion and extension.

Across all six participants, the mean absolute error (MAE) varied only between approximately $30.1^{\circ}$--$32.1^{\circ}$ and $20.8^{\circ}$--$23.7^{\circ}$ without forearm rotation and wrist flexion. Despite the relatively large variation in participant height (158--188 cm), no clear relationship between body height and reconstruction accuracy can be observed.

No consistent trend was observed across the participant group. Within the evaluated range of viewing angles ($0^{\circ}$--$45^{\circ}$), the effect of viewpoint appears to be relatively small compared to the variability introduced by individual poses and reconstructed joints. The left arm generally had a higher MAE than the right arm for most participants and viewpoints, which is likely due to the fact that the turn was made counterclockwise, and the larger the turn, the more the visibility of the left hand was reduced.

In contrast, the performed pose had a substantially stronger influence on estimation accuracy. In particular, poses involving pronounced forearm rotation and palm orientation produced the largest errors, suggesting that self-occlusion and ambiguities in hand orientation remain the dominant sources of reconstruction error.



The computational performance of the vision-based pipeline, excluding physical robot motion execution, was also evaluated. Across ten runs over the complete evaluation dataset, the average processing time was $0.264 \pm 0.001$~s per frame, corresponding to approximately $3.78$ frames per second. MediaPipe Pose and Hands estimation accounted for most of the computational cost, requiring on average $0.154 \pm 0.001$~s and $0.109 \pm 0.001$~s per frame, respectively. In contrast, joint angle reconstruction and mapping required only $0.00112$~s per frame. The maximum observed total processing time was $0.359$~s per frame.

\begin{figure}[H]
  \begin{center}
  \includegraphics[angle=90, width=0.9\textwidth]{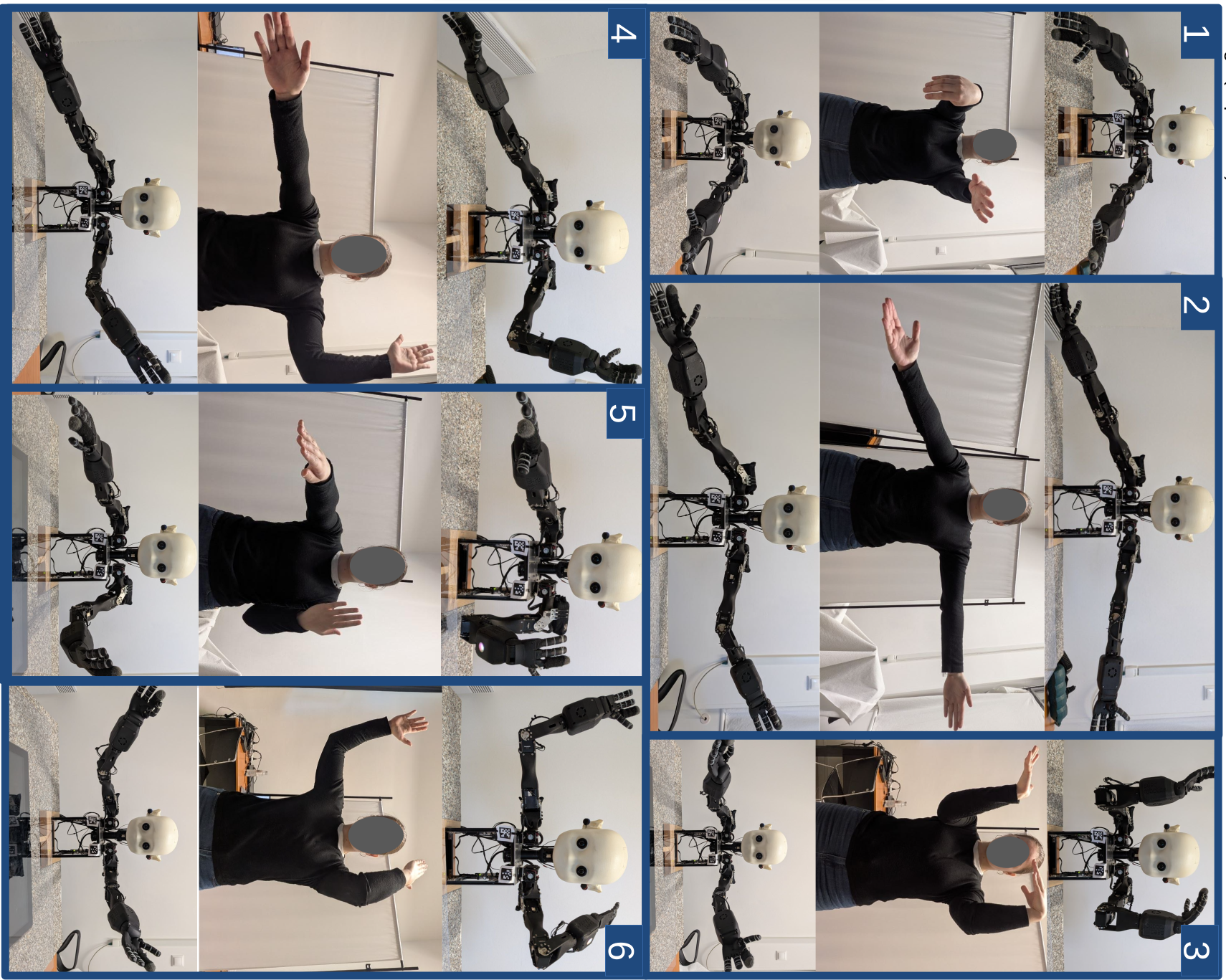} \caption{Qualitative comparison of six test poses. Each cell shows one gesture: the robot pose obtained from manually defined reference landmarks, the corresponding human RGB input image, and the robot pose obtained after MediaPipe landmark extraction and RGB-based reconstruction.}
  \label{fig:experiments-collage}
  \end{center}
\end{figure}

In addition to the quantitative evaluation, Fig.~\ref{fig:experiments-collage} presents qualitative examples of gesture imitation performed by one of the study participants. The gestures shown are different from the predefined evaluation poses and are intended to illustrate the visual behavior of the proposed system on representative arm configurations.

\section{Conclusion}
This paper presented an analytical pipeline for monocular RGB-based imitation of human arm gestures by the semi-humanoid robot NICO. The proposed method combines MediaPipe pose and hand landmark extraction with geometric joint-angle reconstruction and an angle-mapping procedure tailored to NICO's kinematic structure. Unlike learning-based imitation methods, the proposed approach requires no training data and uses a computationally lightweight geometric reconstruction procedure, while providing an interpretable mapping from visual observations to robot motor commands. 
    
Experimental evaluation on a dataset of predefined arm poses demonstrated that the method reconstructs shoulder pitch, elbow bend, and wrist flexion with moderate accuracy, while shoulder roll and especially forearm rotation remain more challenging. The results further showed that incorporating MediaPipe hand landmarks substantially improves forearm rotation estimation compared with using pose landmarks alone, although this joint continues to represent the primary source of reconstruction error. Reconstruction accuracy was relatively consistent across participants and viewing angles under the controlled experimental conditions, whereas the performed gesture had a much greater influence on performance.

The presented method provides a proof of concept for analytical monocular RGB-based gesture imitation using MediaPipe landmarks and geometric reconstruction. While the approach achieves encouraging results for several arm joints under controlled conditions, further improvements and validation on larger and more diverse datasets are required before it can be considered a robust solution for general gesture imitation. Future work will focus on improving forearm rotation estimation, incorporating temporal smoothing, extending the system to dynamic gestures, and evaluating the method on more diverse subjects and viewpoints.

\begin{credits}
\subsubsection{\ackname} 
The authors are thankful to Branislav Zigo for technical assistance.
This research was supported by Horizon Europe MSCA project TRAIL, no. 101072488 and by Slovak Research and Development Agency, project APVV-21-0105.

\subsubsection{\discintname}
The authors have no competing interests to declare that are relevant to the content of this article.

\end{credits}

\end{document}

%% file: figures/pipeline_diagram.tex
\begin{figure}[h]
\centering
\begin{tikzpicture}[
    node distance=0.3cm,
    >=stealth,
    process/.style={
        rectangle,
        rounded corners,
        draw,
        align=center,
        minimum width=3.5cm,
        minimum height=0.9cm,
        font=\small
    },
    data/.style={
    rectangle,
    draw,
    dashed,
    align=center,
    minimum width=3.5cm,
    minimum height=0.9cm,
    font=\small
    },
    arrow/.style={->, thick}
]

\node[data] (input) {RGB input frame};
\node[process, right=of input] (mediapipe) {MediaPipe pose and hand\\landmark extraction};
\node[data, right=of mediapipe] (coords) {3D landmark coordinates\\$P_i = (x_i,y_i,z_i)$};
\node[process, below=of coords] (angles) {Joint angle reconstruction};
\node[process, left=of angles] (mapping) {Angle-to-motor-mapping};
\node[process, left=of mapping] (execution) {Motor command execution\\on NICO robot};

\draw[arrow] (input) -- (mediapipe);
\draw[arrow] (mediapipe) -- (coords);
\draw[arrow] (coords) -- (angles);
\draw[arrow] (angles) -- (mapping);
\draw[arrow] (mapping) -- (execution);

\end{tikzpicture}
\vspace*{-3mm}
\caption{Processing pipeline of the proposed gesture imitation approach. Dashed boxes denote data representations, solid round boxes denote processing steps.   }
\label{fig:pipeline}
\end{figure}